# An Edge Assisted Robust Smart Traffic Management and Signalling System for Guiding Emergency Vehicles During Peak Hours

Shuvadeep Masanta[*,1], Ramyashree Pramanik[1], Sourav Ghosh[2], Tanmay Bhattacharya[3]

[1] Department of Information Technology, Techno Main Salt Lake, Kolkata, India
[2] ORCID: 0000-0003-1866-1408
[3] Associate Professor, Department of Information Technology, Techno Main Salt Lake, Kolkata, India

**Abstract.** Congestion in traffic is an unavoidable circumstance in many cities of India and other countries. It is an issue of major concern. The steep rise in the number of automobiles on the roads followed by old infrastructure, accidents, pedestrian traffic, and traffic rule violations all add to challenging traffic conditions. Given these poor conditions traffic, there is a critical need for automatically detecting and signaling systems. There are already various technologies that are used for traffic management and signaling systems like video analysis, infrared sensors, and wireless sensors. The main issue with these methods is they are very costly and high maintenance is required. In this paper, we have proposed a three-phase system that can guide emergency vehicles and manage traffic based on the degree of congestion. In the first phase, the system processes the captured images and calculates the Index value which is used to discover the degree of congestion. The Index value of a particular road depends on its width and the length up to which the camera captures images of that road. We have to take input for the parameters (length and width) while setting up the system. In the second phase, the system checks whether there are any emergency vehicles present or not in any lane. In the third phase, the whole processing and decision-making part is performed at the edge server. The proposed model is robust and it takes into consideration adverse weather conditions such as hazy, foggy and windy. It works very efficiently in low light conditions also. The edge server is a strategically placed server that provides us with low latency and better connectivity. Using Edge technology in this traffic management system reduces the strain on cloud servers and the system becomes more reliable in real-time because the latency and bandwidth get reduced due to processing at the intermediate edge server.

# 1 Introduction

To increase road safety and manage traffic smoothly, traffic light systems have a major role in traffic safety. Many cities in India face hardship due to the traffic jam that affects the transportation network of the city and causes major dilemmas. The problem of traffic congestion becomes more complex due to the involvement of multiple factors. First of all, traffic flows differently at different point of the day, with its peak usually during the office hours (9:00 am-11:00 am). Weekends reveal minimum load while Monday to Friday shows dense traffic in cities. Secondly the existing traffic light signaling system works on algorithms based on hard-coded delays where the transition time of traffic lights are fixed and does not depend on real-time traffic condition. One more crucial issue is related to emergency vehicles of high priorities like ambulances, and fire brigades that could get stuck in the traffic jam. So, we need to upgrade our conventional traffic signaling system by allowing traffic lights to signal in such a way that will be based on the congestion level and give passage to the emergency vehicles a higher priority. Thus, allowing the traffic to be smooth.

# 2 Related Work

Several researches were reported in the area of intelligent traffic light control systems, most of them focused on sensing technology, communication, and decision making. A research work focusses on building a distributed framework whose primary purpose is to categorize the congestion level of urban road network in real-time. This particular framework uses VANET(Vehicular ad-hoc network) to find out the positions of vehicle on road, it implements temporal and spatial methods on data from a case study. [1]. An intelligent traffic system has been introduced which works on real-time. It uses Big data and IoT to offer better accuracy by deploying traffic indicators in order to reevaluate the traffic related data for each instance of events occurring in the system. In an interval of 500 meters and 1000 meters sensors are installed in the roads. These sensors have the ability to detect vehicle on the road. These sensors collect data related to the traffic condition which is then transferred to the big data analytics center for processing. The density of traffic is calculated by analyzing the processed data with intelligent tools and solutions are defined accordingly. [2]. The technological advancement of cities relies hugely on Mobile Cloud Computing System. A Cloud-based Smart healthcare application have been developed using mobile cloud computing and Big data. The Page Rank algorithm has been used to perform a comparative analysis and study on Apache graph and Hama's graph. [3]. Many other methods have been suggested which controls the signalling time on the real time traffic loads like SCOOT [4], RHODES [5], Sydney Cooperative Adaptive Traffic System (SCATS) [6], Green Link Determining (GLIDE) [7],[8] etc. Inductive sensors placed near traffic junction collects data and then this data is used for deciding the timer limit. Wireless communication is required between centralized server and the sensors for effective output.[10][11]. Due to transmission of huge amount of data the system becomes slow. A paper suggests a set of fuzzy rules that are made to use data gathered (arrival flow, queue length, exit flow) from road sensors and calculate the amount of time for the next phase for improving the traffic flow on an isolated intersection.

[12] This method is very much time consuming and condition for each area is different. This paper has pointed out on the issue of traffic congestion in urban areas [18] by suggesting a method which optimizes the traffic light as a solution. SUMO simulator and a PSO optimization technique are used in this approach for traffic lights cycle .[13] [19] Simulation-Based Vehicular Traffic Lights Optimization. The main issue in this method is delay and there is no special solution for emergency vehicles. A finite-interval model has been proposed in a paper that finds optimal green time for each of the three phases of light timing cycle. The goal is to obtain a satisfactory solution, with the help of Bat algorithm (BA) to reduce wastage time at a junction.[14]

The paper proposed processing of real-time image to build a smart traffic controller. Images will be captured in sequences. Digital image processing applications are used to analyze these image sequence to detect vehicle and according to the analyzed results the road traffic lights are controlled. 19]

## 3   Methodology

The main problem with the previous methods is that they are very costly and high maintenance is required and the existing traffic light signaling system works on algorithms based on hard-coded delays where the transition time of traffic lights are fixed and does not depend on real-time traffic condition.

Our suggested three phases system for smart traffic management and signaling operates on the incoming traffic on the basis of traffic congestion. The proposed model for detecting traffic congestion is primarily based on two factors, the first factor is calculating the vehicle count from the captured images and the second factor takes into account the presence of Emergency vehicles (EV).

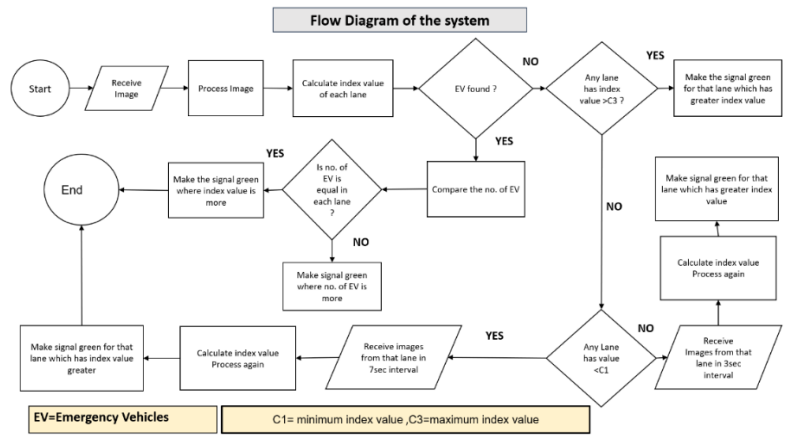

**Fig. 1.** Workflow of smart traffic signal system

### 3.1 Calculating Index value

In the first phase photos captured from cameras at the signal are used as input in our suggested algorithms which are processed using YOLO and the total number of vehicles is calculated.

Based on the statistical studies of the day-to-day traffic trends, an index value has to be calculated from the count of vehicles present at a moment in each lane. This index value calculation helps us to understand the congestion level of traffic and it also determines the priority levels for each lane.

As per our algorithm, there is a maximum threshold time, after which the signal light will change from red to green even if the congestion level is below the minimum index. This threshold time will be counted from the last time the signal was turned red from green. After the threshold time expires the priority for that particular lane increases.

The index values have three predetermined levels C1, C2, and C3.

i) If the calculated index is less than C1(minimum congestion level) the signal light will remain red until the maximum threshold time expires, as it will have the lowest priority. If other lanes also have index values less than C1 then which lane has a greater index value, signal light of that lane will be green. pictures from this lane will be clicked at a span of 7sec.

ii) If the calculated index is more than C3(maximum congestion level) the signal light will turn green for that lane. If other lanes also have index values more than C3 then which lane has a greater index value, the signal light of that lane will be green.

iii) If the calculated index is C2 which is between C1 and C3, there will be no signal light change but images will be captured at an interval of 3sec.

We are comparing traffic congestion on each side of the road by using index values. This value helps us to make the decision of which side light should be green. The index value differs in every area according to the traffic flow and road size in every area. To calculate the index value first we count the number of cars in the signal using YOLOV3. Then we divide the number of vehicles by the size of the area of a certain part of the road.

Suppose at a signal we scanned and count the number of vehicles, the number is 'x'. The width of the road is 'd' and we consider the length of the road 'L'.

So, the index value will be

$$Cx = x/d*L$$

For each area, we specify the three parameters' index values (C1, C2, C3) according to the traffic flow of the area. Then calculate the index value for real-time traffic and compare the value with the three parameters according to which the decision is made.

**Algorithm 1. Calculate the index value and compare the congestion**

**Input:** Captured image

**Output:** Signal phase

1. Degree_of_congestion () {
2. **read** image
3. **read** C1, C2, C3
4. calculate index _value
5. **if** (index_value <= $C_1$)
   **Then**
       Signal_phase red
6. **end if**
7. **if** (index_value >= $C_3$)
   **Then**
       Signal_phase green
8. **end if**
9. **if** (index_value>C1 && index_value<C3)
   **Then**
       Go to step 2
10. **return** signal phase}

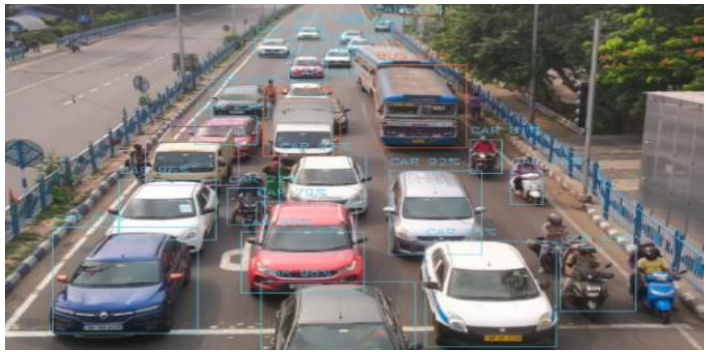

**Fig. 2.** Counting the number of vehicles.

### 3.2 Arrival of Emergency Vehicle

To deal with the arrival of the emergency vehicles toward the intersection of road, an image processing algorithm (including OCR) is applied to the captured images to identify an ambulance or fire bridge. To identify an ambulance the images are transformed to mirror images and Optical Character Recognition (OCR) is done on those images to search for "AMBULANCE" text on the image. For fire-bridged OCR is applied on the original image. If an emergency vehicle is detected in any lane, then the main priority goes to that lane and the signal for that lane will be made green while. other lanes will be turned red.

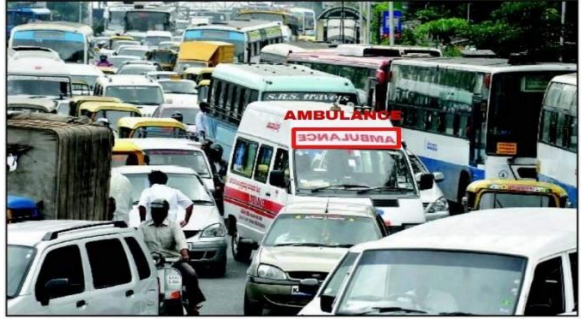

**Fig. 3.** Detection of Ambulance

1. If more than one lane has an emergency vehicle then which lane has a greater number of emergency vehicles signal of that lane will be green.
   If lane one has 1 Emergency Vehicle and lane three has 3 Emergency Vehicle,
   $NV_1 < NV_3$ [$NV_1$=number of emergency vehicle in lane 1, $NV_3$= number of emergency vehicle in lane 3]
   Then, the signal of lane 3 will be green.
I. If the number of emergency vehicles is equal in lanes, then which lane has a greater index value than that lane, signal of that lane will be green, and the other will be red.

### 3.3 Decision making at Edge server

To reduce latency between image processing and reflect outcome we are using the Edge server to process the image and decision making. By using the Edge server, the processing and decision making is done locally there is very less latency between input and output. The images captured by all the cameras connected at the intersection will be processed within edge server, this will be repeated for each side of the road to identify traffic conditions, and change the traffic lights dynamically according to situation.

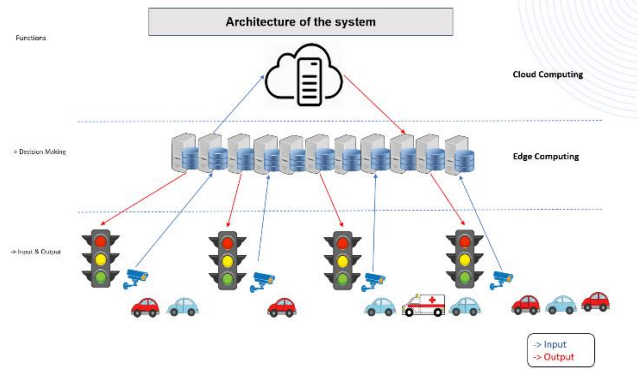

**Fig. 4.** Suggest architecture of server and application in traffic lighting system

Table 2. Algorithm of Decision Making

**Input:** Image

**Output:** Signal_phase

   **Step 1:** Start
   **Step 2:** Let $NV_1$, $NV_2$, $NV_3$, and $NV_4$ number of emergency vehicles in each lane
   **Step 3: Read** Images from each lane
   **Step 4:** Store $NV_i$ in int amb_count []     // i from 0 to 4
   **Step 5: If** ($NV_i$ != 0)             // i from 0 to 4
       Go to step 7
   **Step 6: If** ($NV_i$=0)              // i from 0 to 4
       Call Degree_of_congestion
   **Step 7:** calculate the max of amb_count and store all indexes in a list     //index of max value is saved in a list
   **Step 8: if** (count(list)==1)
       Index=List [0], make signal_phase of index lane green
   **Step 9: if** (count(list)>1)
       Call Degree_of_congestion for each index in the list
   **Step 10: End**

For uploading and downloading any configuration of data the edge servers are required to be connected to the cloud. However, for rest of the work i.e., for processing the data, connection with the cloud is not necessary. So, the system can also work without internet.

Since the processing and analyzing of the images are done on the edge devices itself, so we don't need to upload all the images to the cloud, this result in lesser data being sent to the cloud, resulting in efficient utilization of the network bandwidth. Therefore, our smart traffic light control system will be bandwidth-efficient and can be applied in remote areas.

## 4 Experiment Results for different Weather conditions

Index Value > 0.25 (Green Light)

Index Value < 0.10 (Red Light)

0.10 < Index Value < 0.25 (Wait)

**Table 3.** Result of various weather condition

| Weather condition | System Working | Result (Index value) | output |
|---|---|---|---|
| Clear Sunny Weather | YES | **0.26 (Green light)** | 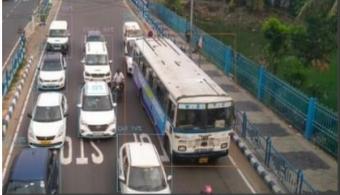 |
| Rainy weather | YES | **0.3 (Green Light)** | 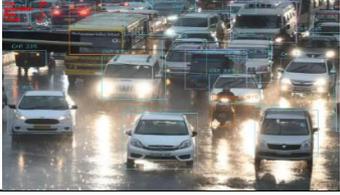 |
| Heavy snowfall | YES | **0.12 (wait)** | 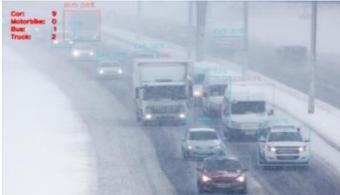 |
| Foggy weather | YES | **0.04 (Red Light)** | 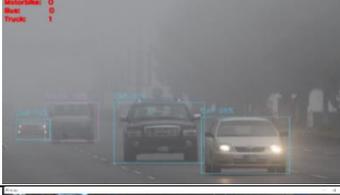 |
| Rainy and Night | YES | **0.13 (Wait)** | 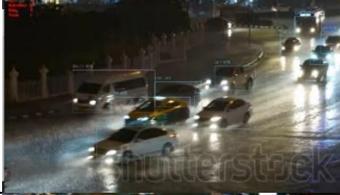 |
| Night | YES | **0.2 (wait)** | 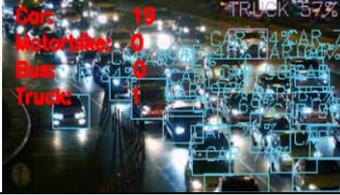 |

# 5   Conclusion

To reduce traffic congestion which is a major problem faced in major cities today, we have proposed a three-phase smart automatic Traffic Management System. We combine YOLO object detection, Tesseract, and Mobile Edge Technology to develop this traffic management system. The object detection model was built to analyse images and give outputs the total number of vehicles in the images. An index value is calculated from the total count of vehicles present at a moment in each lane. While we calculate the index value we take into consideration of length and width of the road. This calculated index value helps us to understand the congestion level of traffic and it also determines the priority levels for each lane by comparing with the pre-set threshold value (this pre-set value changes according to the length and width of the road). According to the index value, our algorithm will decide what should be traffic colour of the lane. As this system consider the length and width of the road that's why it is robust and it can be applied for any road. Tesseract help to detect emergency vehicles using optical character recognition. By calculating the number of emergency vehicles on each lane and by comparing their number the system decides which lane should be green. As we test the system for different weather conditions every time it gives accurate result. This system is independent of weather it can work in almost any weather condition.  Edge technology is being used in this traffic management system so as to reduce the strain on cloud servers. The decision making on the basis of index value and numbers of emergency vehicles is performed in edge server. As edge server is located near the system so the system is more reliable in real-time because the latency and bandwidth get reduced due to processing at the intermediate edge server. So, this model is cost-efficient and its time complexity is less.